\documentclass[letterpaper]{article} 
\usepackage{aaai2026}  
\usepackage{times}  
\usepackage{helvet}  
\usepackage{courier}  
\usepackage[hyphens]{url}  
\usepackage{graphicx} 
\usepackage{natbib}  
\usepackage{caption} 
\usepackage{algorithm}
\usepackage{algorithmic}

\usepackage{newfloat}
\usepackage{listings}
\DeclareCaptionStyle{ruled}{labelfont=normalfont,labelsep=colon,strut=off} 
\floatstyle{ruled}
\newfloat{listing}{tb}{lst}{}
\floatname{listing}{Listing}

\title{Enhancing Image Aesthetics with Dual-Conditioned Diffusion Models Guided by Multimodal Perception}
\author{
    Xinyu Nan\textsuperscript{\rm 1} \quad Ning Wang\textsuperscript{\rm 2} \quad Yuyao Zhai\textsuperscript{\rm 1} \quad Mei Yang\textsuperscript{\rm 1*}
}
\affiliations{
    \textsuperscript{\rm 1}Peking University, Beijing, China\\

    \textsuperscript{\rm 2}University of Science and Technology of China, Anhui, China\\
}

\usepackage{bibentry}

\begin{document}

\maketitle

\begin{abstract}
Image aesthetic enhancement aims to perceive aesthetic deficiencies in images and perform corresponding editing operations, which is highly challenging and requires the model to possess creativity and aesthetic perception capabilities. Although recent advancements in image editing models have significantly enhanced their controllability and flexibility, they struggle with enhancing image aesthetic. The primary challenges are twofold: first, following editing instructions with aesthetic perception is difficult, and second, there is a scarcity of ``perfectly-paired'' images that have consistent content but distinct aesthetic qualities. In this paper, we propose Dual-supervised Image Aesthetic Enhancement (DIAE), a diffusion-based generative model with multimodal aesthetic perception. First, DIAE incorporates Multimodal Aesthetic Perception (MAP) to convert the ambiguous aesthetic instruction into explicit guidance by (i) employing detailed, standardized aesthetic instructions across multiple aesthetic attributes, and (ii) utilizing multimodal control signals derived from text-image pairs that maintain consistency within the same aesthetic attribute. Second, to mitigate the lack of ``perfectly-paired'' images, we collect ``imperfectly-paired'' dataset called IIAEData, consisting of images with varying aesthetic qualities while sharing identical semantics. To better leverage the weak matching characteristics of IIAEData during training, a dual-branch supervision framework is also introduced for weakly supervised image aesthetic enhancement.
Experimental results demonstrate that DIAE outperforms the baselines and obtains superior image aesthetic scores and image content consistency scores. 
\end{abstract}


\section{Introduction}

Image aesthetic enhancement (IAE) is a valuable visual application of creative artificial intelligence (creative AI), signifying that AI possesses human-like aesthetic perception capabilities to create more attractive works in fields like fashion, advertising and art~\cite{imagereward,kirstain2023pickapicopendatasetuser,wu2023human}. IAE requires models to be capable of both image aesthetic assessment and image attributes enhancement, recognizing low quality attributes such as color, composition, and texture, and adjusting them to high quality~\cite{IAQAReview}.

\begin{figure}
    \centering
    \includegraphics[width=\linewidth]{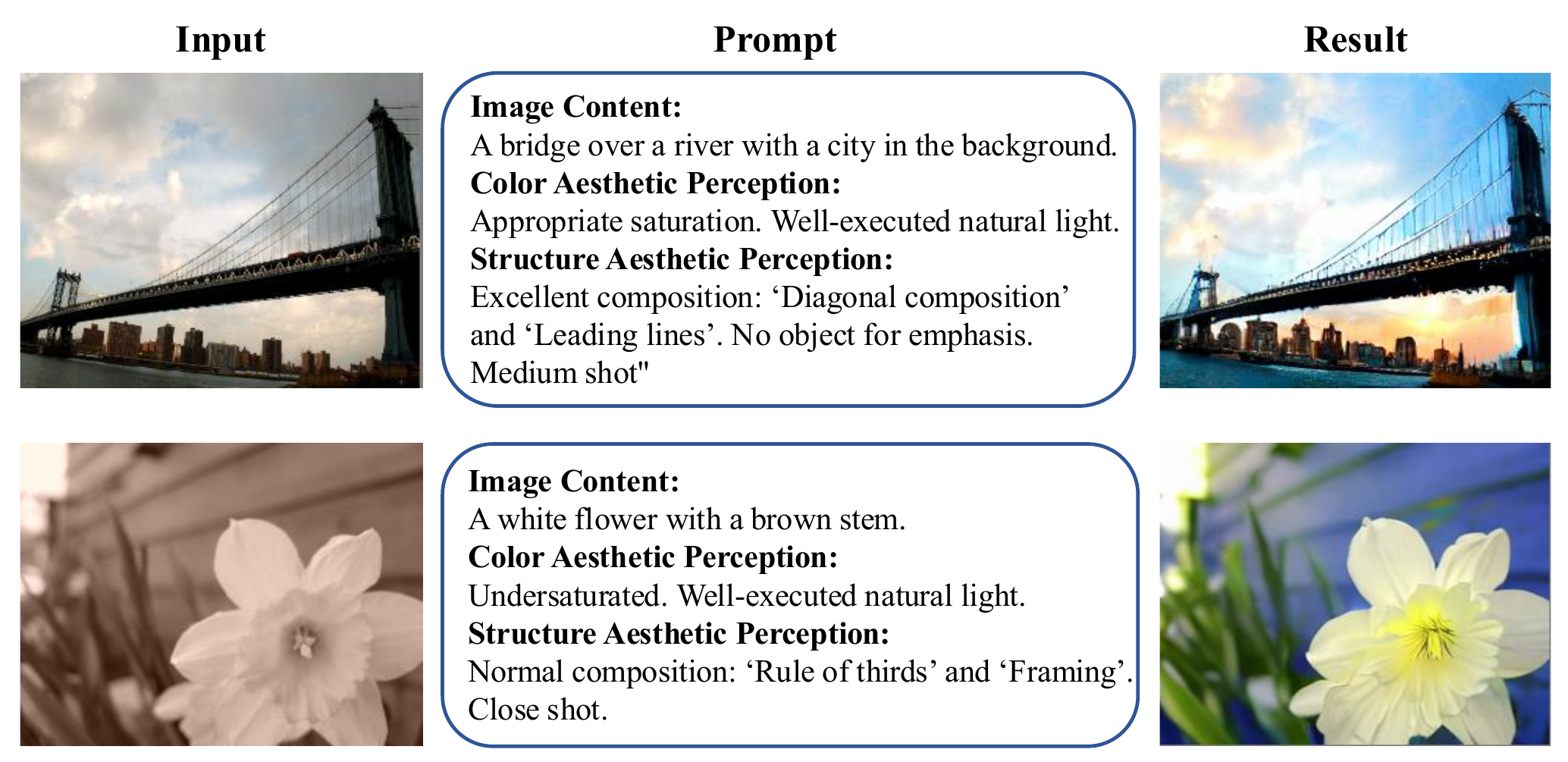}
    \caption{
    Examples of image aesthetic enhancement. Given the original image, image content and aesthetic descriptions (left), DIAE generates the results with enhanced aesthetic (right). Our DIAE is capable of generating images that are content-consistent with input images while possessing enhanced aesthetics.}
    \label{fig0:intro}
\end{figure}

\begin{figure}
    \centering
    \includegraphics[width=\linewidth]{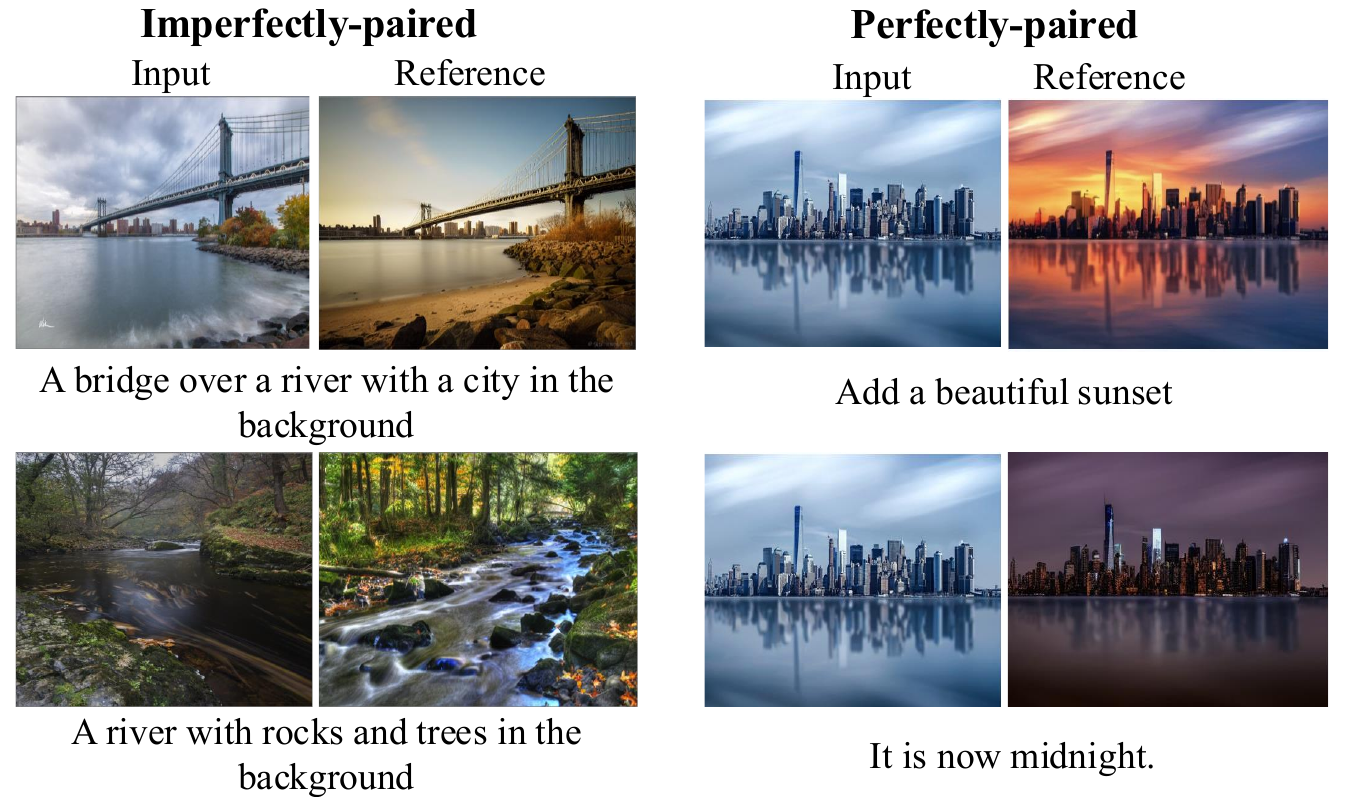}
    \caption{Comparison between ``imperfectly-paired' and
    ``perfectly-paired''~\protect\cite{brooks2022instructpix2pix}}
    \label{fig1:imperfectly}
\end{figure}

EnhanceGAN~\cite{EnhanceGAN}, based on GAN~\cite{GAN}, was one of the first attempts to use generative models for Image Aesthetic Enhancement (IAE). EnhanceGAN simplifies image attributes enhancement into image color enhancement and image cropping modules within a \textit{Generator}, and relies on a \textit{Discriminator} to evaluate the aesthetic quality of generated results. With the significant success of diffusion-based models in various image generation tasks, their robust semantic generation capabilities and diverse outputs offer more extensible image aesthetic assessment and attributes enhancement compared to GAN-based models. Among current diffusion-based generative methods, image editing methods~\cite{SmartBrush,brooks2022instructpix2pix,dragdiffusion,li2023stylediffusion} can achieve IAE to some extent. However, existing diffusion-based image editing methods face two significant challenges in addressing IAE: 1) Aesthetic perception is a high-level human visual capability, influenced by many uncontrollable elements, such as cultural background, personal experiences, and emotional states. This makes it difficult for the text encoder alone to comprehend and effectively guide the generation direction of diffusion-based models; 2) Generative models require ``perfectly-paired'' images that differ only in edited regions for fully supervised training, shown in~\ref{fig1:imperfectly}. For image aesthetic enhancement tasks, these ``perfectly-paired'' images must show enhancement solely in aesthetic attributes, which is extremely expensive to obtain as it requires professional annotations from artists.

To tackle above challenges, we propose Dual-supervised Image Aesthetic Enhancement (DIAE), a novel image aesthetic enhancement approach based on diffusion model. To simulate the aesthetic perception of generative models, we introduce Multimodal Aesthetic Perception (MAP), a language-visual combined instruction comprehension module. MAP processes formatted textual aesthetic assessment alongside corresponding visual representations. Specifically, we follow commonly used aesthetic analysis standard and categorize aesthetic assessment into image color and image structure~\cite{EnhanceGAN}. Their corresponding visual representations are different due to their different aesthetic attributes. The visual representations for color attributes are HSV maps, which align with human color perception, while structure attributes are represented by contour maps that highlight object outlines and spatial arrangements within the scene. Moreover, inspired by the success of ControlNet~\cite{zhang2023adding} in style editing, MAP incorporates the visual representations alongside textual aesthetic assessment into the diffusion model as control signals, allowing the generative model to perform comprehensive aesthetic enhancements.

To address the lack of ``perfectly-paired'' data for fully supervision, we propose to use ``imperfectly paired'' data for weak supervision to train generative models.
Specifically, we introduce a new dataset called Imperfectly-paired Image Aesthetic Enhancement Data (IIAEData). Each triplet in the dataset consists of ``imperfectly-paired'' input and reference images, along with an image caption and aesthetic assessment. The ``imperfectly-paired'' images are constricted with same semantics, but their structure, style and art are not constricted, as Fig.~\ref{fig1:imperfectly} shown. Due to the weak matching of ``imperfectly-paired'' images, directly using the reference image, which is inconsistent with the content of the input, as the sole supervision is not feasible. Instead, we introduce a dual-branch supervision framework for model optimization, which divides the training process of the diffusion model into a semantic supervision branch supervised by the input image, and an aesthetic supervision branch supervised by the reference image. With IIAEData and the dual-branch supervision framework, the image aesthetic enhancement task is treated as weakly-supervised diffusion-based generation. Based on the model optimization method, DIAE is able to generate images that preserve the content of the input image while aligning the aesthetic attributes with those of the reference image, shown in Fig.~\ref{fig0:intro}.

We conduct comprehensive experiments to validate DIAE across multiple dimensions, including aesthetic quality~\cite{LAION-Aesthetics-v1,LAION-Aesthetics-v2}and content consistency~\cite{hessel2022clipscorereferencefreeevaluationmetric}. Our results demonstrate that DIAE can efficiently enhance image aesthetics while ensuring content consistency and generate results that better align with human aesthetic preferences compared to SOTA image editing methods. Furthermore, our proposed DIAE can be seamlessly integrated with MLLMs to generate aesthetic assessments for MAP. This enables DIAE to function as an end-to-end aesthetic enhancement model, directly catering to a diverse range of user preferences. In summary, our contributions are threefold:

\begin{itemize}
    \item We introduce an image editing method with Multimodal Aesthetic Enhancement (MAP), called Dual-supervised Image Aesthetic Enhancement (DIAE). MAP provides multimodal aesthetic perception for DIAE.
    \item We introduce the ``Imperfectly-paired'' Image Aesthetic Enhancement (IIAE) dataset and a dual-branch supervision framework for training DIAE to achieve image aesthetic enhancement. This addresses the lack of image pairs with same semantics but different aesthetics. 
    \item Our DIAE achieves image aesthetic enhancement efficiently and exhibits superior performance compared to existing image editing methods in terms of aesthetic.
\end{itemize}

\section{Related Work}

\textbf{Image aesthetic assessment.} Image Aesthetic Assessment (IAA) in computer vision science is to evaluate photography techniques and artistic approaches of images~\cite{IAA}. Early works in the field of IAA focused on designing score-based IAA models to efficiently and accurately predict the Mean Opinion Score (MOS), exploring how human aesthetic judgment can be translated into computational learning processes. For instance, works like TAVAR~\cite{TAVAR} and NIMA~\cite{2017NIMA} proposed a CNN-based reasoning approach to simulate human perceptual processes in image aesthetics. He et al.~\cite{TADNet} introduced a novel dataset TAD66K and a theme-aesthetic network that adaptively learns to predict aesthetic rules based on identified themes. As large language models (LLMs) have demonstrated remarkable understanding and generative capabilities, recent studies incorporate LLMs into IAA. For example, Q-ALIGN~\cite{wu2023qalign} trained language models using discrete scoring levels defined by textual descriptions rather than direct scores, effectively simulating human subjective rating processes. Furthermore, several works~\cite{zhou2024uniaa,depictqa_v1,depictqa_v2, 2025MonetGPT} leveraged MLLMs for IAA, generating textual descriptions and scores for image aesthetics.

\textbf{Image aesthetic enhancement.} Early works~\cite{2018Exposure,2020NICER,EnhanceGAN} focused on unifying various image enhancement methods, such as color enhancement, contrast improvement, and image cropping, to achieve image aesthetic enhancement.
Although this works achieve image content consistency well, their enhancement effects are quite limited. 
Recent several Text-to-Image (T2I) and Image-to-Image (I2I) approaches have attempted to use aesthetic scores or human aesthetic preference to guide the image generation process. DOODL~\cite{wallace2023end} and RAHF~\cite{RAHF} utilize MOS of the image as classifier guidance during the diffusion model generation process. By adjusting the gradients of the image classifier~\cite{wallace2023end}, they effectively steer the aesthetic generation of images for both I2I and T2I. Additionally, DiffusionDPO~\cite{wallace2023diffusionmodelalignmentusing} and MPS~\cite{zhang2024MPS} leverage reinforcement learning to train T2I diffusion models that incorporate human preferences, guiding models to generate images that better align with human aesthetic preferences. 

\textbf{Diffusion-based generative models.} Apart from T2I, diffusion-based models have also demonstrated impressive performance in the fields of image editing, image style transfer, and image enhancement. 
Existing diffusion-based~\cite{SD1,SD2} works have introduced hundreds of methods, like a text and shape guided object inpainting diffusion model\cite{SmartBrush}, a diffusion model without per-example fine-tuning or inversion~\cite{brooks2022instructpix2pix} and a text transformation query, automatically finds a ROI mask covering the image region to be edited~\cite{couairon2022diffedit} for semantic editing, a training-free and prompt-free diffusion model~\cite{parmar2023zeroshotimagetoimagetranslation}, one-shot stylization method that is robust in structure preservation~\cite{One-Shot-Style} and pure content and style feature fusion network based on content feature extractor and style feature extractor~\cite{Puff-Net} for style editing and an interactive point-based editing with diffusion models~\cite{dragdiffusion} and image inpainting methods~\cite{LatentPaint,lugmayr2022repaintinpaintingusingdenoising,kim2024radregionawarediffusionmodels} for structure editing. However, none of these image editing methods are concerned about the image aesthetic by editing the aesthetic attributes of the images, such as color, light and composition. 

\begin{figure*}
    \centering
    \includegraphics[width=\textwidth]{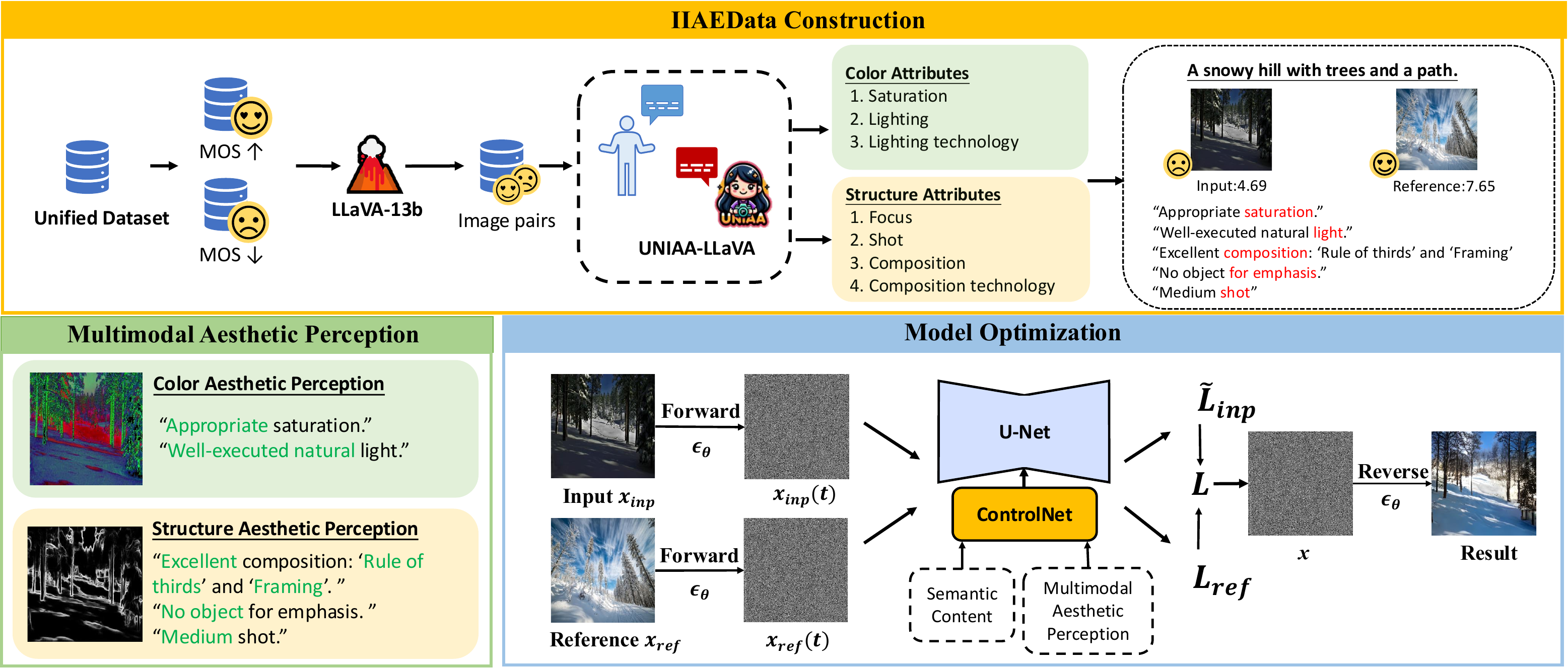}
    \caption{Overview of DIAE.
    \textbf{IIAEData Collection}: collecting `imperfectly-paired'' data for IIAEData, obtaining the image pairs matching and aesthetic assessment prompts through LLaVA-13b~\cite{liu2023llava} and UNIAA-LLaVA~\cite{zhou2024uniaa}.
    \textbf{Multimodal Aesthetic Perception (MAP)}: multimodal aesthetic perception with textual descriptions and HSV and contour maps for image color and image structure.
    \textbf{Model Optimization}: weakly-supervised diffusion model training strategy with `imperfectly-paired'' input and reference image, while using MAP through ControlNet~\cite{zhang2023adding}.}
    \label{fig2:overview}
\end{figure*}

\section{Method}
\subsection{IIAEData: ``Imperfectly-paired'' Dataset}
Currently, the data triplet in public datasets related to aesthetics is generally as single image with a Mean Opinion Score (MOS) annotation and an image content classification label, without matched paired images that can be directly used for training fully-supervised diffusion-based generative models. Constructing image pairs for image aesthetic enhancement not only requires time-consuming manual editing of each low-quality image but also demands that the editors be experienced artists or photographers, making the construction cost unaffordable. 
Another approach is to refer to image quality assessment (IQV), most works in IQV obtain image pairs with completely consistent image content but differences in image quality by artificially degrading high-quality images, but such degradations typically reflect image quality rather than aesthetic~\cite{zhou2022quality,thong2022contentdiverse,ghildyal2022stlpips}. Image aesthetic focuses more on whether the image is pleasing and artistic, which is incredible to simulate through artificial degradations. Therefore, artificial degradations are inappropriate for image aesthetic enhancement.
Based on this situation, we propose a dataset entirely composed of weakly matched data pairs, called ``imperfectly-paired''(shown in Fig.~\ref{fig1:imperfectly}), combined with a weakly supervised training strategy to train a diffusion model for image aesthetic enhancement.

\textbf{Data Collection.} As illustrated in Fig.~\ref{fig2:overview}, we begin by collecting several aesthetic datasets, AVA~\cite{AVA}, TADNet~\cite{TADNet}, KonIQ~\cite{KONIQ}, and FLICKR~\cite{FLICRK}, to build a unified dataset. These datasets comprise images captured under natural conditions rather than generated. This means that any distortions present in the images, such as blurriness and noise, reflect real-world scenarios and natural aesthetics. Each image in the datasets is labeled with a MOS. 
To clearly distinguish between the aesthetic qualities of input and reference images, we select images with high MOS values as reference image candidates and those with low MOS values as input image candidates, deliberately excluding images with intermediate scores.

\textbf{Formulating ``Imperfectly-paired'' Image Pairs.} Consequently, we apply LLaVA-13b~\cite{liu2023llava} to generate an image caption of each image. Based on these captions, we match images to obtain pairs with the same semantics, which are ``imperfectly-paired'' and the comparison between ``perfectly-paired'' and ``imperfectly-paired'' images is shown in Fig.~\ref{fig1:imperfectly}. The image in the pair with lower MOS is input image and the other is reference image. Subsequently, we employ UNIAA-LLaVA~\cite{zhou2024uniaa}, a MLLM with strong capability of image aesthetic assessment, to generate textual aesthetic assessments of the input images in terms of color, lighting, lighting techniques, composition, composition techniques, focus and shot types. The textual aesthetic assessments are then standardized to a specific format, which is helpful for diffusion models to comprehend. Finally, we have invited human experts to manually review the generated captions and prompts and remove incorrect or disputed cases.

\subsection{Multimodal Aesthetic Perception}
Although we can obtain accurate image aesthetic assessments using aesthetic MLLMs, these assessments are often highly abstract and confusing, like \textit{``Oversaturated with warm tone palette.''}, and \textit{```Rule of thirds' composition method and `Framing' composition technique.''}. Furthermore, the simple text encoders of diffusion models are almost incapable of understanding the aesthetic assessment instructions to generate images with enhanced aesthetics. In other words, current diffusion models lack aesthetic perception capabilities. Multimodal Aesthetic Perception (MAP) is proposed to effectively enable diffusion models to gain aesthetic perception with a lightweight and efficient structure based on ControlNet~\cite{zhang2023adding}, shown in Fig.~\ref{fig2:overview}. 

 \textbf{Constructing Aesthetic Perception.} Considering that image color and image structure are widely acknowledged as critical attributes that effectively reflect image aesthetics, and their ease of quantitative assessment makes them prevalent in image aesthetic researches in computer vision~\cite{IAQAReview}, the image aesthetic assessments in IIAEData are firstly categorized into two parts: image color and image structure. Each part is composed of crucial aesthetic attributes, elements and techniques. The color attributes include saturation, lighting and lighting techniques. The structure attributes include focus, shot types, composition and composition techniques. However, all of those assessments are ambiguous and difficult for diffusion models to understand. 

 \textbf{Adding Visual Perception.} In order to comprehend and follow the textual ambiguous aesthetic assessments, we introduce another modality to enrich the aesthetic perception. For color attributes, we introduce HSV maps to reflect the visual representations of saturation, lighting and lighting techniques, because HSV maps contain hue, saturation and value of images that align with the human perception of color attributions of images. Compared to RGB, it can more intuitively express the details of colors. We utilize contour maps generated by HED models~\cite{hed} for structure attributes, which emphasize the focus, shot, composition and composition techniques of images. Meanwhile, HSV maps and contour maps respectively explore the intuitive visual representation of the color space and the two-dimensional space of the image. Compared to RGB, they lose some semantic information to a certain extent. Therefore, we add color and spatial textual information of the image as a language modality to assist the model in enhancing its aesthetic understanding capability. Therefore, the aesthetic perception is finally represented as the textual descriptions combined with corresponding visual representations for color and structure attributes, which employ visual representations to support diffusion models in understanding ambiguous aesthetic assessments. 

\begin{figure}[h]
    \centering
    \includegraphics[width=\linewidth]{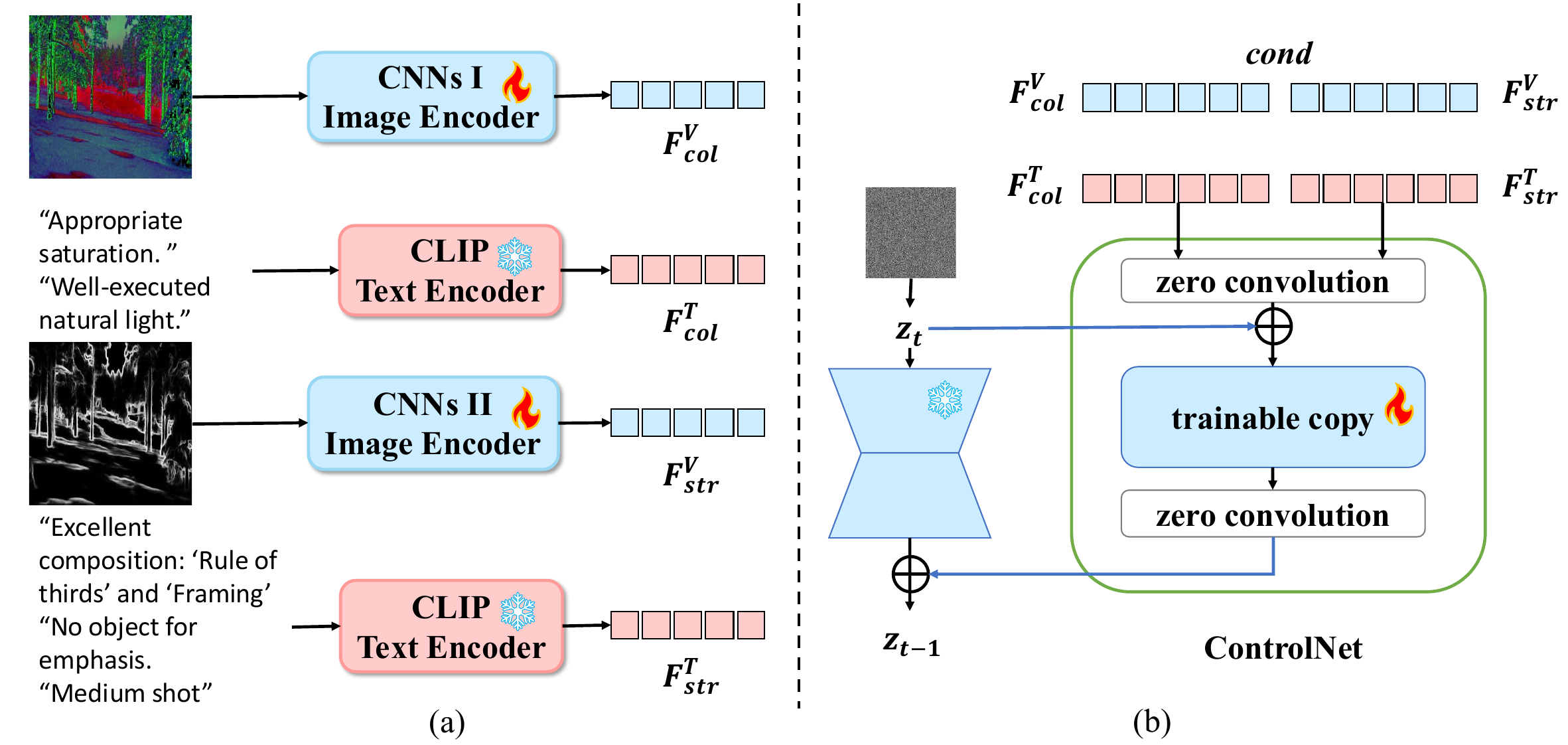}
    \caption{Guiding DIAE via MAP. (a) Conditioning embeddings $cond$ generation. (b) Using $cond$ to guide diffusion models via ControlNet~\protect\cite{zhang2023adding}.}
    \label{fig3:aesthetic-perception}
\end{figure}

\textbf{Guiding generation with MAP.} To guide diffusion models with MAP, we adopt an adding structure to introduce additional control signals, inspired by the architecture of ControlNet~\cite{zhang2023adding}. Specifically, we obtain the control signals, which indicate the color and structure of images, for diffusion models. The control signals consist of visual and textual conditioning embeddings $(F^{I}_{col}, F^{T}_{str})$ and $(F^{I}_{col}, F^{T}_{str})$. As Fig.~\ref{fig3:aesthetic-perception}(a) illustrated, we train two branches based on convolutional neural networks (CNNs) $\Phi_{i}$ to extract image features from HSV maps $I_{col}$ and contour maps $I_{str}$ as visual conditioning embeddings, $F^{I}_{col}$ and $F^{I}_{str}$. We also utilize the well-trained CLIP text encoder~\cite{clip} $\Phi_{t}$ for textual conditioning embeddings, $T_{col}$ and $T_{str}$:
\begin{equation}
    F^{I}_{col} = \Phi_{i}(I_{col}),F^{I}_{str} = \Phi_{i}(I_{str})
\end{equation}
\begin{equation}
    F^{T}_{col} = \Phi_{t}(T_{col}),F^{T}_{str} = \Phi_{t}(T_{str})
\end{equation}
As Fig.~\ref{fig3:aesthetic-perception}(b) shown, the control signals are then applied in the UNet of the diffusion model, allowing the injection of aesthetic perception into the diffusion model and controlling the formation of aesthetic attributes during the generation process.
The control signals $cond$ consist of $\{cond_{h},cond_{c}\}$ for controlling the diffusion model, which can be written as:
\begin{equation}
    \{cond_{h},cond_{c}\} = \{(F^{I}_{col}, F^{T}_{col}),(F^{I}_{col}, F^{T}_{str})\}
\end{equation}

MAP is a novel multimodal fusion module, which combines the aesthetic descriptions and the corresponding visual representations and guides image editing via an adding structure for diffusion models. With the guide of MAP, diffusion models can comprehend ambiguous aesthetic instructions and perform practical editing operations effectively. 

\subsection{Model Optimization}

Since the reference image and input image are "imperfectly-paired" in terms of semantic consistency, a novel model optimization method is essential to ensure that the results generated by the diffusion models are "perfectly-paired" with the input images in terms of content, while only enhancing the aesthetics.
Inspired by the C-S disentangled framework used in StyleDiffusion~\cite{li2023stylediffusion}, we introduce a dual-branch supervision framework to DIAE.

\textbf{Modifying the Denoising Training.}
The image generation of diffusion models involves starting from pure noise and gradually removing noise to generate the image. The training objective is to minimize denoising error, which is critical for diffusion models to learn how to effectively remove noise at each step. The dual-branch supervision framework modifies this denoising process by using both the input image and the reference image for supervision. For the timestep-wise denoising of diffusion models, early denoising timesteps concentrate on constructing the semantics of the output result, while creating aesthetic attributes such as color, lighting, and focus in later denoising timesteps. 

\textbf{Dual-branch Supervision Framework.}
In the dual-branch supervision framework, we divide the training process into two periods via a parameter $t_{s}$. The training of denoising is supervised by the input image in the triplet when timestep $t\leq t_{s}$, while supervised by the reference image in all the timesteps. The loss functions for the supervision of the input image and reference image are $L_{inp}$ and $L_{ref}$, respectively.  
The loss $L$ for DIAE is shown below:
\begin{equation}
    L = L_{ref} + \lambda L_{inp}
\end{equation}
\begin{equation}
    L_{ref} = \Vert \epsilon_{ref} - \epsilon_{\theta}(x_{ref}(t), c, x_{ref}, t, cond) \Vert_2^2
\end{equation}
\begin{equation}
    L_{inp} = \Vert \epsilon_{inp} - \epsilon_{\theta}(x_{inp}(t\%t_{s}), c, x_{inp}, t\%t_{s}, cond) \Vert_2^2
\end{equation}
where $x_{ref},x_{inp}$ are reference and input images, $\epsilon_{ref},\epsilon_{inp}$ are noise for reference image and input image. $\lambda$ is the weight coefficient to balance $L_{ref}$ and $L_{inp}$.

In the dual-branch supervision framework, $t\_{s}$ is utilized to control the optimization objective of the diffusion model. $L\_{inp}$ aims to supervise the initial stages ($t\%t\_{s}$, for timestep $\leq$ $t\_{s}$) to keep semantics of the input image unaffected and prevent the aesthetic attributes of the input image from interfering with the generation in later timesteps. 
While generating aesthetic attributes image when timestep $\textgreater t\_{s}$ is fully supervised by the high MOS reference, $L_{ref}$.

\section{Experiments}
\subsection{Experiments Settings}
\textbf{Implementation Details.}
We utilize pre-trained stable diffusion-v1.5 (SD-v1.5) as the diffusion model. Compared to the latest Stable-Diffusion 3.5, SD-v1.5 has fewer parameters and relatively stable performance. It can achieve satisfactory generative effects while ensuring high generative efficiency. SD-v1.5 comprises a CLIP text encoder to encode textual prompts, an autoencoder to compress images into latent space, a UNet to predict the noise at every timestep and other structures. We adopt ControlNet~\cite{zhang2023adding} as an extension to enhance the controllability of SD-v1.5, guiding SD-v1.5 by using additional conditional inputs beyond text. Additionally, we modify the initial architecture of ControlNet~\cite{zhang2023adding} by using two CNN-based image encoders to encode HSV and contour maps of the input image. The parameters of UNet and ControlNet~\cite{zhang2023adding} are trainable and the parameters of CLIP text encoder are frozen, while $t_{s}$ is default set as 900. We adopt AdamW optimizer with a learning rate of 1e-5 and train on 4 NVIDIA A800 GPUs for 100,000 iterations. We generate the results at the resolutions of $256\times256$ and $512\times512$, which take about 4 and 9 seconds with 50 denoising steps, respectively

\begin{table*}[ht]
    \centering
    \begin{tabular}{c c c c c c c c c}
    \hline
               & \multicolumn{4}{c}{\textbf{Aesthetics Metrics}} & \multicolumn{2}{c}{\textbf{Content Consistency Metrics}} \\
        Methods& \multicolumn{2}{c}{\textbf{LAIONs-based Score}} & \multicolumn{2}{c}{\textbf{MLLMs-based Score}} & \multicolumn{2}{c}{\textbf{CLIP-I Score}} \\ 
               & (res = 256) & (res = 512) & (res = 256) & (res = 512) & (res = 256) & (res = 512) \\
    \hline
       Original image  & 4.962 & 5.123 & 3.243 & 3.300 & 1.000 & 1.000 \\
       ControlNet      & 4.979 & 5.522 & 3.271 & 3.415 & 0.628 & 0.617 \\
       InstructPix2Pix & 4.991 & 5.396 & 3.264 & 3.325 & 0.764 & 0.690 \\
       MGIE            & 4.947 & 5.519 & 3.045 & 3.411 & 0.557 & 0.770 \\
       DOODL           & 5.102 & 5.140 & 3.255 & 3.297 & 0.775 & 0.703 \\
       DIAE            & 5.324 & 6.012 & 3.339 & 3.662 & 0.772 & 0.784 \\
    \hline
    \end{tabular}
    \caption{Quantitative comparison with the state-of-the-arts. LAIONs-based Score and MLLMs-based Score are scored by LAIONS~\protect\cite{LAION-Aesthetics-v1,LAION-Aesthetics-v2,LAION-Aesthetics-v2.5} and MLLMs~\protect\cite{zhou2024uniaa,depictqa_v2}, respectively.}
    \label{tab1:Overall results}
\end{table*}

\textbf{Data.}
Considering the significant differences in aesthetic analysis between portraits and other categories of images, it is not suitable for processing as a general aesthetic enhancement task. Specifically, we have deleted the training samples containing highly personalized semantic content such as ``portrait'' and ``people''. Ultimately, We collected 47.5K samples from AVA~\cite{AVA}, TAD66K~\cite{TADNet}, KONIQ~\cite{KONIQ} and FLICRK~\cite{FLICRK}, consisting of 45K training and 1.5K testing samples. Each sample contains a input image and a reference image, a caption (brief overview of image content) and an textual aesthetic assessment of the input image. 

\textbf{Evaluation.}
We evaluate the performance of the generation models in terms of image aesthetic quality and image content consistency.
For image aesthetic quality assessment, we employ the mainstream aesthetic scoring models, LAION-Aesthetic Predictors~\cite{LAION-Aesthetics-v1,LAION-Aesthetics-v2,LAION-Aesthetics-v2.5} to score the results (LAIONs-based Score: the average score of testing data, which is scaled to (0,10)). We also adopt MLLMs, UNIAA-LLaVA~\cite{zhou2024uniaa} and DepictQA-v2~\cite{depictqa_v2} for scoring the results (MLLMs-based Score: the average score of testing data, which is scaled to (0,5)). 
For image content consistency, we utilize the CLIP-I score~\cite{hessel2022clipscorereferencefreeevaluationmetric} to evaluate the similarity between the result image and the input image.




\begin{figure}[h]
    \centering
    \includegraphics[width=\linewidth]{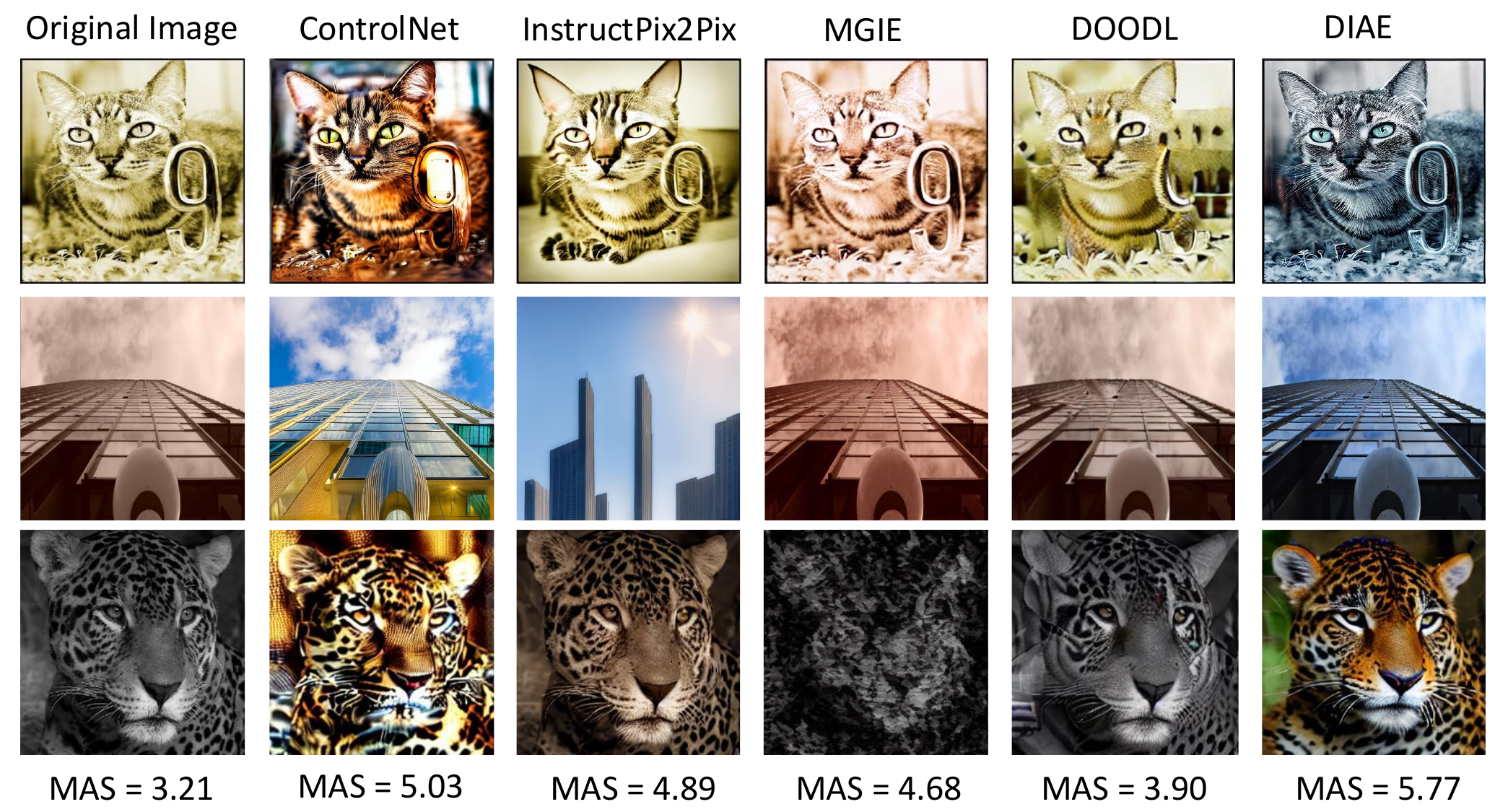}
    \caption{Examples of generative results from low aesthetics quality batch (MOS $<4.0$). The examples with resolution of $512\times512$, while the Mean Aesthetic Score (MAS) below is the LAIONs-based Score calculated on the batch.}
    \label{fig6:comparison and ablation}
\end{figure}

\subsection{Comparison with State-of-the-Arts}
\begin{figure}[h]
    \centering
    \includegraphics[width=\linewidth]{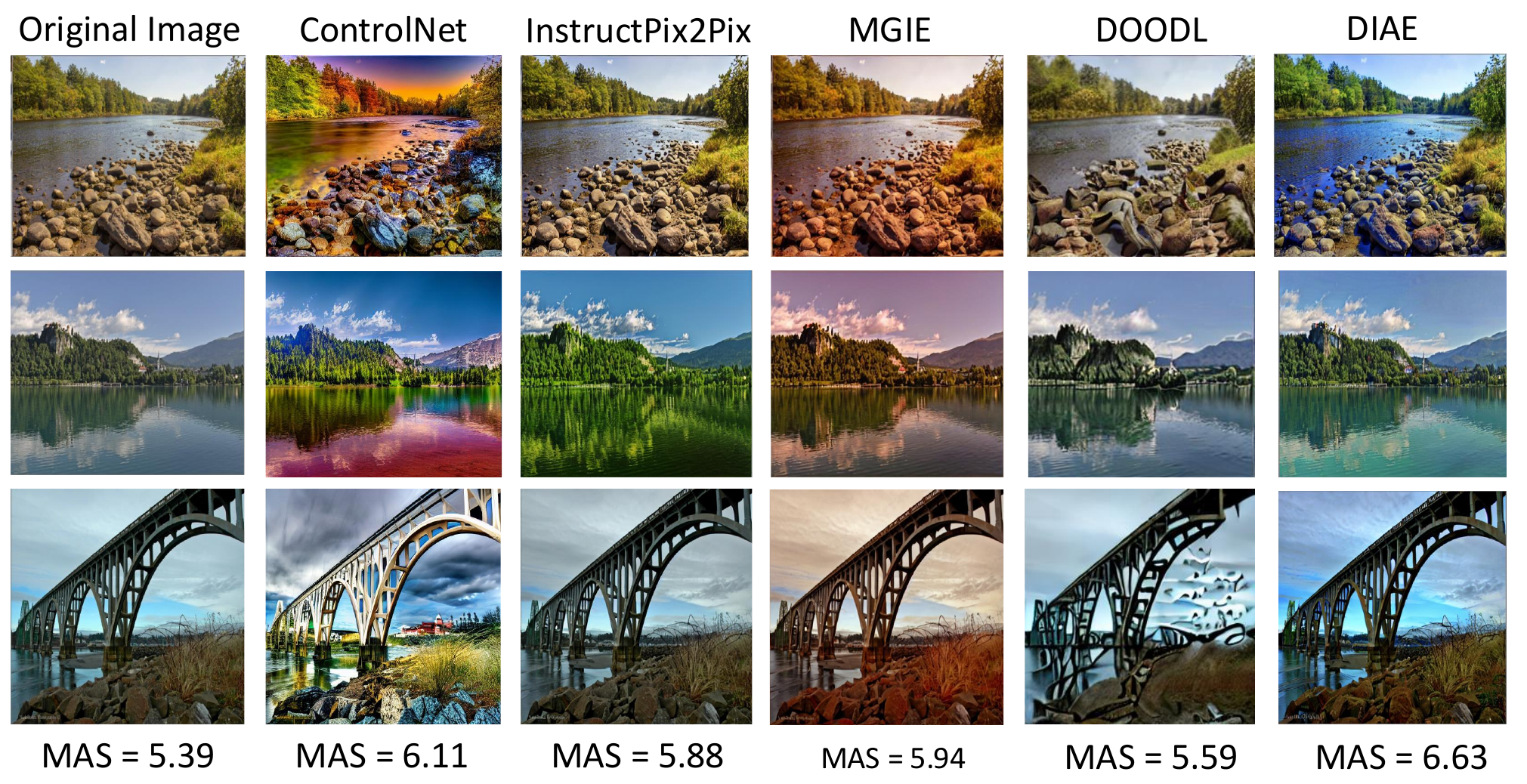}
    \caption{Examples of generative results from high aesthetics quality batch (MOS $>5.0$). The examples with resolution of $512\times512$, while the Mean Aesthetic Score (MAS) below is the LAIONs-based Score calculated on the batch.}
    \label{fig7:comparison and ablation}
\end{figure}

We collect different SOTA diffusion-based image editing methods using Stable Diffusion v1.5 as diffusion backbone, ControlNet~\cite{zhang2023adding}, InstructPix2Pix~\cite{brooks2022instructpix2pix}, MGIE~\cite{fu2024mgie}, DOODL~\cite{wallace2023end} as baselines. We offer different suitable prompts for the SOTA methods to achieve image aesthetic enhancements (shown in Appendix).
We also discussed in more detail the performance of DIAE under different original image qualities. We divided the testing images into two batches: one batch represents low aesthetic quality samples with LAIONs-based Mean Aesthetic Score (MAS) below 4.0, and the other batch represents high aesthetic quality samples with MAS above 5.0. We selected some samples from the two batches, as shown in Fig.~\ref{fig6:comparison and ablation} and Fig.~\ref{fig7:comparison and ablation}, and calculated the MAS of the results generated by different methods on these two batches.

\textbf{Image Aesthetics Comparison.}
As Tab.~\ref{tab1:Overall results} shown, DIAE achieve the highest aesthetic scores in both the LAIONs-based Scores and Aesthetics-MLLMs Scores. For $256\times256$ images, DIAE's LAIONs-based Score and MLLMs-based Score improved by 7.3\% and 3.0\%, respectively.
For $512\times512$ images, DIAE's LAIONs-based Scores and MLLMs-based Scores improved by 17.4\% and 11.0\%, respectively.

As shown in Fig.~\ref{fig6:comparison and ablation}, for images with very poor aesthetic quality (MAS$ < 4.0$), these images have obvious color and brightness defects. Additionally, images with no color intensity variation lead to a lack of the targets of images. The images in Fig.~\ref{fig6:comparison and ablation} obtain the image aesthetic assessments from MLLM (UNIAA~\cite{zhou2024uniaa}), such as \textit{``Undersaturation''}, \textit{``Poor light''}, and \textit{``No target in the image.''}. We performed inference on these images using DIAE and other SOTA methods. It can be seen that compared to other methods, DIAE shows significant advantages in image color and brightness, and the noticeable variation in color intensity further achieves image re-targeting.

As shown in the Fig.~\ref{fig7:comparison and ablation}, for images with slightly better aesthetic quality (MAS$ > 5.0$), these images do not have obvious defects in color, brightness, composition, etc. Such images require further optimization of color contrast and brightness contrast to make the focus of the image more prominent and beautified. We also performed inference on these images using DIAE and other SOTA methods. It can be seen that compared to other methods, DIAE achieves a higher increase in MOS scores, with better preservation of original image details and higher color authenticity.

\textbf{Image Content Consistency Comparison.} The image content consistency is scored by CLIP-I Score~\cite{hessel2022clipscorereferencefreeevaluationmetric}, which can measure the semantic similarity between the input image and the generated result.
As Fig.~\ref{tab1:Overall results} shown, the CLIP scores~\cite{hessel2022clipscorereferencefreeevaluationmetric} of DIAE achieved 0.772 and 0.784 on $256\times256$ and $512\times512$ images, respectively, which outperform the other methods. 
As Fig.~\ref{fig6:comparison and ablation} and Fig.~\ref{fig7:comparison and ablation} demonstrated, our DIAE shows advantages in preserving the content details of the original image compared to other methods. Unlike other methods (such as the first-row \textit{cat} image in Fig.~\ref{fig6:comparison and ablation} by other methods), our DIAE dose not add or remove details from the original image. Our method also does not create something out of nothing (such as InstructPix2Pix~\cite{brooks2022instructpix2pix} generating completely unrelated buildings in the second row of Fig.~\ref{fig6:comparison and ablation}).

\subsection{Ablation Study}
\begin{figure}[h]
    \centering
    \includegraphics[width=\linewidth]{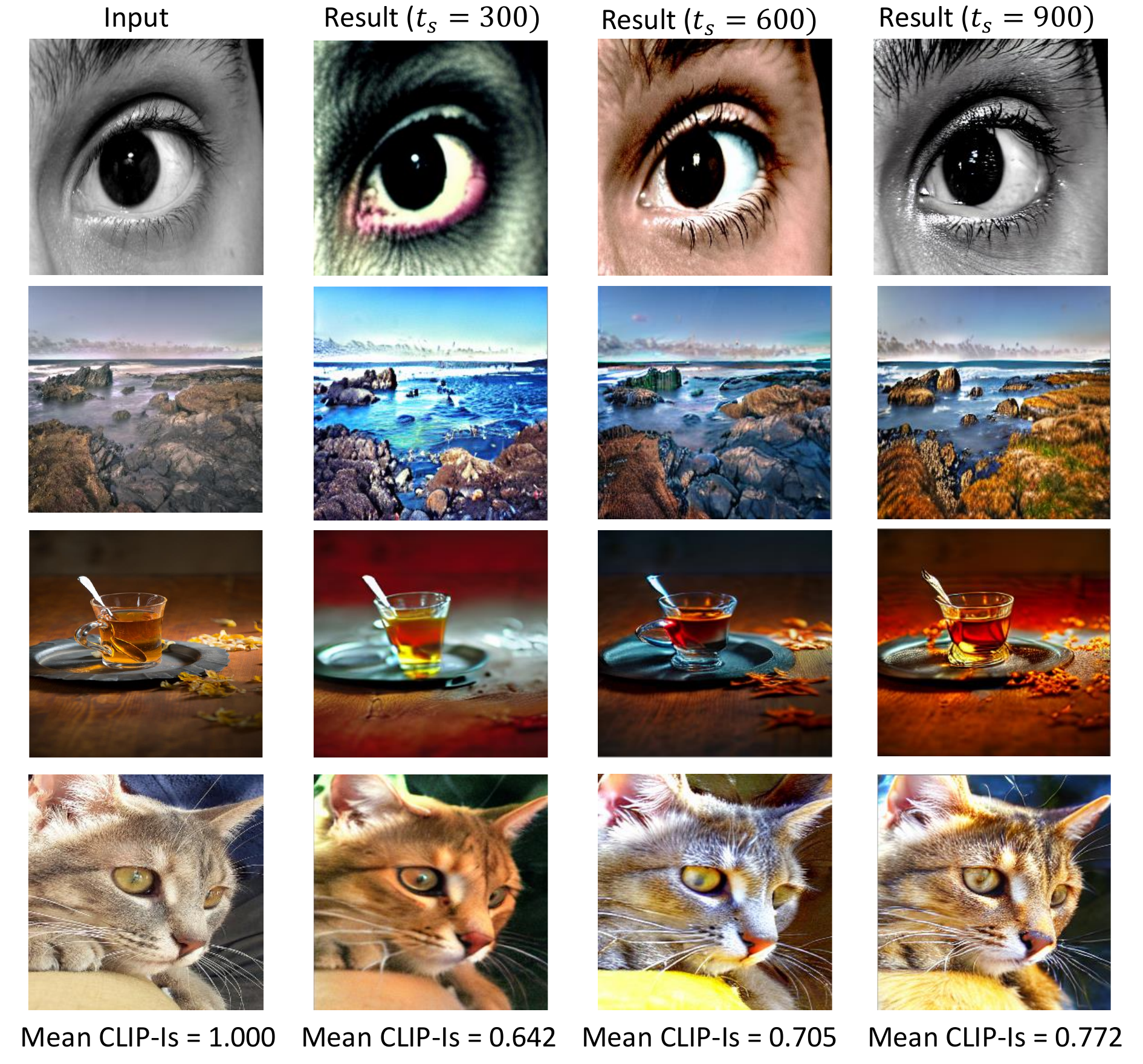}
    \caption{Comparison between DIAE with different $t_{s}$. CLIP-I: the similarity of generated and input images.}
    \label{fig5: the value of ts}
\end{figure}

\textbf{Study on the value of $t_{s}$.} 
The parameter $t_s$ determines how much the DIAE's generated image guided by the input image or the reference image. Specifically, a smaller $t_s$ increases supervision by the reference image during more denoising timesteps, while a larger $t_s$ reduces this supervision, allowing greater focus on the input image. We have conducted extensive experiments to evaluate the impact of $t_{s}$ on generated results. As Fig.~\ref{fig5: the value of ts} demonstrated, as $t_{s}$ increases, the generated images retain more aesthetic elements, such as color and brightness, from the input image(Note that the total number of timesteps is 1000). We also computed the mean CLIP-I Scores of different generated results based on different $t_{s}$.
These results indicate the larger $t_{s}$ indicates rentaining more details of input images, and adjusting $t_{s}$ influences the aesthetic attributes of generated images. 

\textbf{Impact of Multimodal aesthetic perception.}
We evaluate the impact of MAP on DIAE by withdrawing the visual modality guidance and textual modality guidance from MAP separately. When the visual modality is withdrawed from MAP and a text prompt completely consistent with MAP is provided as guidance (requiring a contour map, as it is a necessary input for ControlNet), DIAE degrades to ControlNet~\cite{zhang2023adding} in the comparative experiment, called DIAE ($-w/o$ $v$). As shown in tab.~\ref{tab2:ablation of map}, its performance in aesthetic enhancement and original content preservation is inferior to DIAE. When the text modality is withdrawed from MAP and HSV and contour maps completely consistent with MAP are provided as guidance (the text prompt only includes the image caption, i.e., content semantics, without any aesthetic text evaluation), called DIAE ($-w/o$ $t$). As shown in tab.~\ref{tab2:ablation of map}, its aesthetic enhancement is also inferior to DIAE. However, it still has advantages over other methods, especially in content consistency, indicating that introducing dual visual modalities as guidance further enhances the model's advantage in content consistency of the generated results.

\begin{table}[h]
    \centering
    \begin{tabular}{c c c c}
    \hline
       Methods & LAIONs & MLLMs & CLIP-I\\
    \hline
       DIAE ($-w/o$ $v$)       & 5.250 & 3.343 & 0.623 \\
       DIAE ($-w/o$ $t$)       & 5.428 & 3.410 & 0.792 \\
       DIAE                   & 5.668 & 3.501 & 0.778 \\
    \hline
    \end{tabular}
    \caption{Ablation study with DIAE, DIAE ($-w/o$ $v$) and DIAE ($-w/o$ $t$) to evaluate the impact of multimodal aesthetic perception. The LAIONs, MLLMs and CLIP-I indicate the $256\times256$ and $512\times512$ images' mean LAIONs-based, MLLMs-based and CLIP-I Scores, respectively.}
    \label{tab2:ablation of map}
\end{table}
\section{Limitation and Conclusion}
\textbf{Limitation.}
Currently, DIAE performs well in image aesthetic enhancement for scenes such as landscapes, animals, architecture, and still life. However, image aesthetic enhancement for images of portraits and crowds is quite complex, as it involves not only color, brightness, and composition, but also facial features and body shape, which are important factors affecting viewers' aesthetic perception. Since there is a lack of systematic research on related aesthetic assessment, we did not construct such data in DIAE's training data, and DIAE has not yet been developed for image aesthetic enhancement in portrait-related scenes.

\textbf{Conclusion.}
In this paper, we analyze the challenges faced by diffusion models in the aesthetic quality enhancement task and subsequently propose DIAE. DIAE is based on the diffusion-model and leverages MAP to comprehend aesthetic instructions. Based on the weakly aligned dataset we constructed, we introduce a dual supervision framework for training DIAE. Experiments demonstrate that DIAE effectively enhances the aesthetic quality of images while maintaining the consistency of generated image content. Human users can employ DIAE to enhance the aesthetics of any image by providing their own aesthetic instructions or by utilizing aesthetic comprehension MLLM to generate such instructions. This work represents an initial exploration of diffusion models in aesthetic tasks. Future work can expand the volume of training data and refine instruction optimization to make a profound study.

\bibliography{aaai2026}

\end{document}